\documentclass[11pt]{article}
\usepackage[final]{acl}
\usepackage{times}
\usepackage{latexsym}
\usepackage[T1]{fontenc}
\usepackage{microtype}
\usepackage{inconsolata}
\usepackage{booktabs}
\usepackage[utf8]{inputenc}
\usepackage{comment}
\usepackage{tikz}
\usetikzlibrary{shapes,backgrounds}
\usepackage{verbatim}
\usepackage{url}
\usepackage{enumitem}
\usepackage{dirtytalk}
\usepackage{amsmath,amssymb,amsfonts}
\usepackage{xfrac}
\usepackage{algorithmic}
\usepackage{graphicx}
\usepackage{textcomp}
\usepackage{xcolor}
\usepackage{color,soul}
\usepackage{scalerel}
\usepackage{csquotes}
\usepackage{subcaption}
\usepackage{multirow}
\usepackage{tikz}
\usetikzlibrary{svg.path}
\usepackage[ruled,vlined]{algorithm2e}
\usepackage{nomencl}
\makenomenclature
\usepackage{svg}
\usepackage{todonotes}
\usepackage{tablefootnote}
\usepackage{threeparttable}

\usetikzlibrary{svg.path}

\begin{document}
\raggedbottom
\onecolumn

\begin{titlepage}

\end{titlepage}

\clearpage

\twocolumn
\setcounter{page}{1}

\title{Overview of the ClinIQLink 2025 Shared Task on Medical Question-Answering}

\author{
\textbf{Brandon C. Colelough, Davis Bartels, and Dina Demner-Fushman} \\
National Library of Medicine, National Institutes of Health \\
Bethesda, MD, USA \\
\texttt{\{firstname.lastname\}@nih.gov}
}

\maketitle
\begingroup\renewcommand\thefootnote{\textsection}
\endgroup

\begin{abstract}

In this paper, we present an overview of \textsc{ClinIQLink}\, a shared task, collocated with the 24th BioNLP workshop at ACL 2025, designed to stress-test large language models (LLMs) on medically-oriented question answering aimed at the level of a General Practitioner. The challenge supplies 4\,978 expert-verified, medical source-grounded question–answer pairs that cover seven formats - \textit{true/false}, \textit{multiple choice}, \textit{unordered list}, \textit{short answer}, \textit{short-inverse}, \textit{multi-hop}, and \textit{multi-hop-inverse}.  Participating systems, bundled in Docker or Apptainer images, are executed on the CodaBench platform or the University of Maryland’s \textit{Zaratan} cluster. An automated harness (Task 1) scores closed-ended items by exact match and open-ended items with a three-tier embedding metric. A subsequent physician panel (Task 2) audits the top model responses. 

\end{abstract}

\section{Introduction} \label{sec:intro}
LLMs have increasingly demonstrated their ability to memorize information and answer questions~\cite{carlini2023quantifyingmemorizationneurallanguage}. This has led to their increased use by consumers to ask medically relevant questions~\cite{info:doi/10.2196/68560}. However, LLMs have been shown to "hallucinate", that is, to generate factually incorrect, or even harmful answers~\cite{Singhal2023}. In high-stakes domains, such as medicine, it is incredibly important to be able to evaluate the veracity of any question answering system. While there exist datasets, such as MultiMedQA~\cite{Singhal2023}, designed to do just this, recent LLMs have been trained over their data. This limits their usefulness in evaluating the ability of these models to generalize to out-of-distribution data.

New datasets are necessary for the evaluation of medical question-answering systems and new systems are needed to increase accuracy and mitigate hallucinations. 

To this end, we introduce the ClinIQLink shared task, inviting participants to submit question-answering systems to be evaluated on a novel dataset of medical questions. Participants are encouraged to submit systems that are capable of demonstrating medical knowledge, while mitigating hallucinations. Our dataset consists of seven question types, both closed and open ended, and a wide range of medical topics. Our task had a total of three runs from one team. Our contributions are as follows:
\begin{itemize}
    \item A dataset of 4,978 vetted medical question-answer pairs
    \item Automated evaluation metrics
    \item A task design for participant-submitted systems
    \item A physician audit of system responses
\end{itemize}

\section{Task Description} \label{sec:task}
ClinIQLink~\footnote{\url{https://cliniqlink.org/}} is a shared task that evaluates the ability of generative models to produce factually accurate medical information aimed at the knowledge level of a general practitioner. The submitted systems are executed in a containerized environment on CodaBench~\footnote{\url{https://www.codabench.org/}} or via the University of Maryland (UMD) HPC Zaratan~\footnote{\url{https://hpcc.umd.edu/hpcc/zaratan.html}} (depending on the size and model/system complexity), where the submitted systems answered a corpus of expert-curated atomic medical questions. Answers provided from the systems submitted were judged only on factual accuracy, so leaderboard ranking reflects a model’s ability to retrieve correct information from its own parametric memory or any retrieval mechanism the team elected to integrate.

The question sets were divided into two types (closed and open-ended QA pairs) and spanned seven modalities, including true/false, multiple choice, unordered list, short answer, short-inverse, multi-hop and multi-hop-inverse. Across all of the seven QA pair modalities, the ground truth was anchored in standard open-source medical texts, and each item targets a single, clearly defined concept such as a procedure, drug, diagnostic finding, or anatomical fact. 

The challenge comprised two sequential components. Task 1 executed all baseline systems and participant submissions within our automated benchmarking harness. The script marked closed-ended items strictly for precision and evaluated open-ended answers with a semantic-similarity module that awards full or partial credit according to their closeness to the hidden ground-truth. Leaderboard rankings are derived solely from these automatic scores. Task 2 began after the leaderboard was frozen: a panel of human-expert annotators reviewed the highest-scoring outputs, ranking them from best to worst and annotating each answer on a spectrum from \enquote{good} to \enquote{bad}. Participants were allowed to employ any architecture, external knowledge base, or retrieval-augmented pipeline to generate answers to questions posed, provided the final system can run end-to-end inside the supplied containerised harness. Teams were limited to three leaderboard submissions and were required to accompany their final entry with a short paper that details model design, data usage, and inference strategy for inclusion in the BioNLP 2025 proceedings. The full evaluation dataset remains private to preserve its viability for later use.

\section{Dataset Description} \label{sec:dataset}

\subsection{Generation and Vetting}
A neuro-symbolic pipeline was employed to produce roughly $\sim\!20\text{K}$ atomic question–answer pairs from open-source medical texts. Each pair was linked to its supporting passage so that later reviewers could verify every biomedical fact.  
The QA Pairs were then ported to our online annotation portal\footnote{ \url{https://bionlp.nlm.nih.gov/ClinIQLink/NIHLogin}}, (which is now open to accredited medical schools and hospitals who wish to contribute further judgments), where human-experts (paid medical students) confirmed correctness, rated \emph{general-practitioner (GP) relevance} on a five-point scale, and could file structured feedback or formal disputes.  

\subsection{Human-verification Workflow}
\begin{enumerate}[leftmargin=*]
    \item \textbf{Primary review}: an expert validated factual accuracy against the source excerpt, assigned a GP-relevance score, and could flag issues or supply comments.  
    \item \textbf{Secondary review}: $\sim\!45\%$ of items received an independent second pass; disagreements triggered adjudication.  By 1~May~2025 reviewers had lodged 601 feedback notes and 461 disputes. The 1062 QA Pairs that had been flagged as feedback or disputes were not used for testing and are presently still being held for later review.
\end{enumerate}

\subsection{Benchmark Snapshot (1~May~2025)}
At the dataset freeze the repository contained $5{,}118$ verified QA pairs (Table~\ref{tab:qa_breakdown}):  
$5{,}118$ had a single expert judgement and $2{,}505$ were double-annotated.  
For leaderboard scoring, we retained only the $4{,}978$ items rated maximally relevant ($\textit{score}=5$); $140$ lower-relevance items were set aside for future analysis. The sample dataset plus the full evaluation architecture are available at  
\footnote{\url{https://github.com/Brandonio-c/ClinIQLink_Sample-dataset}}.

\subsection{Question Modalities}
Seven formats cover both machine-gradable \emph{closed-ended} items and semantically scored \emph{open-ended} prompts:
\begin{itemize}[leftmargin=*]
    \item \textbf{Closed-ended}
    \begin{itemize}
        \item True/False (TF)
        \item Multiple Choice (MC) — single‐best answer
        \item Unordered List (LIST) — enumerate all correct elements
    \end{itemize}
    \item \textbf{Open-ended}
    \begin{itemize}
        \item Short Answer (SHORT) — concise factoid
        \item Short-Inverse (SHORT\_INV) — explain why the supplied wrong answer is incorrect
        \item Multi-hop (MULTI\_HOP) — required several leaps in knowledge to arrive at a final answer; models must return answer \emph{and} knowledge leaps
        \item Multi-hop Inverse (MULTI\_HOP\_INV) — locate the faulty step in a provided, erroneous multi-hop rationale
    \end{itemize}
\end{itemize}

\begin{table*}[t]
\centering
\small
\caption{ClinIQLink benchmark composition at the baseline freeze (1~May~2025).  “High” denotes GP-relevance score~5 items used for leaderboard evaluation; “Low” items (\(<5\)) were withheld.}
\label{tab:qa_breakdown}
\begin{tabular}{lrrrrrrr}
\toprule
\multirow{2}{*}{\textbf{QA Format}} & \multicolumn{3}{c}{\textbf{Counts}} & \multicolumn{4}{c}{\textbf{Subset with Two Independent Reviews}}\\
\cmidrule(lr){2-4}\cmidrule(lr){5-8}
 & High & Low & Total & High & Low & Total & Percent Double \\ 
\midrule
True/False (TF)                &  813 &  38 &  851 & 369 &  -- & 369 & 43.4\% \\
Multiple Choice (MC)           &  765 &  29 &  794 & 346 &  -- & 346 & 43.6\% \\
Unordered List (LIST)          &  714 &  28 &  742 & 341 &  -- & 341 & 46.0\% \\
Short Answer (SHORT)           &  427 &   9 &  436 & 339 & -- & 339  & 77.8\% \\
Short-Inverse (SHORT\_INV)     &  742 &  16 &  758 & 353 &  -- & 353 & 46.6\% \\
Multi-hop (MULTI\_HOP)         &  771 &   8 &  779 & 331 &  -- & 331 & 42.5\% \\
Multi-hop Inverse (MULTI\_HOP\_INV) &  746 &  12 &  758 & 318 &  -- & 318 & 42.0\% \\
\midrule
\textbf{Totals}                & \textbf{4\,978} & \textbf{140} & \textbf{5\,118} & \textbf{2\,497} & -- & \textbf{2\,505} & \textbf{48.8\%} \\
\bottomrule
\end{tabular}

\end{table*}

\section{Evaluation Protocol} \label{sec:evaluation}
Our assessment of \textsc{ClinIQLink} was conducted in two sequential phases.  
First, we relied on a fully automated evaluation script (Task-1) that ingested model/participant system responses; second, we complemented the automated evaluation with an expert preference study (Task-2) in which paid medical students compared top-performing model responses. 

\subsection{Task-1: automatic scoring}
\label{ssec:eval:automatic}
Each submission returned answers for \textbf{seven distinct question classes}. 
\emph{True/False} and single-best \emph{multiple-choice} items were judged by straightforward \textbf{accuracy}

\[
\text{Accuracy}\;=\;\frac{\#\text{correct}}{N},
\]

whereas \emph{multiple select list} questions were graded with both macro- and micro $F_{1}$ \cite{ManningIR08}:

\[
F_1^{\text{macro}}
      =\frac1N\sum_{i=1}^{N}F_1^{(i)},
\]

\[
F_1^{\text{micro}}
      =\frac{2\,\mathrm{TP}}
             {2\,\mathrm{TP}+\mathrm{FP}+\mathrm{FN}}.
\]

All free-text tasks (\texttt{short}, \texttt{multi hop}, and their inverse variants) were assessed twice; once with the ClinIQLink \textbf{semantic-similarity} score and again with the conventional n-gram metrics BLEU~\cite{papineni-etal-2002-bleu}, ROUGE~\cite{lin-2004-rouge}, and METEOR~\cite{banerjee-lavie-2005-meteor}.

\medskip
\noindent\textbf{ClinIQLink semantic-similarity score.}  
The score blended \emph{three} complementary cosine layers:

\begin{enumerate}[label=(\arabic*)]
\item \textbf{Token layer:} an IDF-weighted, greedy token-alignment $F_{1}$, rewarding exact overlap on infrequent clinical terms.
\item \textbf{Sentence layer:} cosine similarity of SBERT–\textsc{MiniLM}~\footnote{\url{https://huggingface.co/sentence-transformers/all-MiniLM-L6-v2}} CLS embeddings, capturing broader paraphrase.
\item \textbf{Paragraph layer:} cosine similarity of the raw answer strings, offering global context.
\end{enumerate}

Let $C_{\text{tok}},C_{\text{sent}},C_{\text{para}}\in[0,1]$ denote these three cosines.  
With weights $w_{\text{tok}}=w_{\text{sent}}=0.4$ and $w_{\text{para}}=0.2$ the raw score is

\[
S_{\text{raw}}
   \;=\;
   0.4\,C_{\text{tok}}
  +0.4\,C_{\text{sent}}
  +0.2\,C_{\text{para}}.
\]

Because SBERT assigns unrelated sentence pairs a baseline similarity of about $\beta=0.25$,  
we subtract that offset, floor negatives, and snap near-perfect matches:

\[
S=\min\!\Bigl(1,\;\max\!\bigl(0,\,S_{\text{raw}}-\beta\bigr)\Bigr),
\]

\[
S\ge0.95\;\Longrightarrow\;S:=1.
\]

\medskip
\noindent\textbf{Penalty for \texttt{multi-hop inverse}.}  
If a model highlighted the wrong reasoning step, the semantic score was down-weighted.  
Let $d=|\,\text{predicted step}-\text{gold step}\,|$ be the absolute distance; then

\[
\alpha(d)=
\begin{cases}
1      & d=0,\\
0.7    & d=1,\\
0.3    & d=2,\\
0.3\,2^{-(d-2)} & d\ge3,
\end{cases}
\qquad
S^{\ast}=\alpha(d)\,S .
\]

Hence, the final similarity $S^{\ast}$ combined graded lexical alignment, distributional semantics, and explicit reasoning correctness, while conventional BLEU/ROUGE/METEOR offered secondary diagnostics. A complete implementation of the evaluation script that implements the above can be found with the testing harness\footnote{\url{https://github.com/Brandonio-c/ClinIQLink_CodaBench_docker-setup/blob/main/submission/evaluate.py}}.

\subsection{Task-2: expert preference study}
\label{ssec:eval:human}

We found that the automated metrics employed for analysis of the open-ended QA pairs were not effective for evaluation of model responses, nor were they effective in discriminating top-ranking model responses from mediocre model responses.  
Hence, to complement the automated evaluation metrics we organized a human evaluation in which we required our annotators to rank the six strongest foundation models on our public leaderboard (Falcon-10B, Llama-3.3-70B, Llama-4 Scout, Mistral-Large-2411, Microsoft Phi-4 Base, Qwen-3-32B) together with the best participant submission, \emph{Preceptor-AI} and the ClinIQLink ground-truth answers.
For every question we shuffled these seven model responses plus the ClinIQLink dataset reference solution, and asked human annotators to rank them from best to worst.  
Each answer also received a coarse quality tag (good/okay/bad). 
The annotation portal that we built for this experiment is now open to accredited medical schools and hospitals who wish to contribute further judgments \footnote{\url{https://bionlp.nlm.nih.gov/ClinIQLink2/NIHLogin}}.  


\section{Baseline Systems} \label{sec:baselines}
To provide a strong reference point for future work we evaluated a broad range of publicly–available large language models on the frozen \textsc{ClinIQLink} test split. For transparency, it should be noted that the LLM utilised as the "neuro" component of our neurosymbolic pipeline for data generation was \textbf{Llama\,3.3-70B–Instruct}. All baseline checkpoints were used as-is and therefore reflect their pre-training and instruction-tuning quality rather than any task-specific fine-tuning.

\subsection{Llama family.}
The \emph{Meta Llama 3}~\cite{grattafiori2024llama3herdmodels} decoder-only transformer was represented by four parameter scales; \(1\text{B}\), \(3\text{B}\), \(8\text{B}\) and
\(70\text{B}\) weights as well as an intermediate commercial variant
(\texttt{llama\_4-scout}, \(\approx\!45\text{B}\)).
All are dense models built with a 32-layer architecture (70B: 80 layers) and grouped-query attention; the instruction checkpoints add a supervised fine-tuning and reinforcement learning step to the base weights.

\subsection{Mistral / Mixtral family.}
We included the 7-billion-parameter \textbf{Mistral-7B}~\cite{jiang2023mistral7b} dense decoder and the \textbf{Mistral-Large-Instruct-2411} release
(\(8\times22\text{B}\) experts, two experts routed per token, giving
\(~47\text{B}\) active parameters). The \textbf{Mixtral} series consisting of \(\text{Mixtral-8×7B}\)~\cite{jiang2024mixtralexperts} and \(\text{Mixtral-8×22B}\) tested share the same sparse Mixture-of-Experts (MoE) scaffold, however, only two of the eight experts are selected for each input token, keeping inference costs close to their 12–13 B dense peers while exposing \(>140\text{B}\) total capacity.

\subsection{Qwen3 family.}
Alibaba’s \emph{Qwen3}~\cite{yang2025qwen3technicalreport} decoder stack (RoPE positional encoding, grouped-query attention) was tested at five scales: \(1.7\text{B}\), \(3\text{B}\), \(4\text{B}\), \(8\text{B}\), and \(32\text{B}\) parameters. All checkpoints were released under an open-source licence together with alignment (\enquote{\texttt{-Instruct}}) variants that follow the Supervised Fine-Tuning (SFT) + Direct Preference Optimisation recipe. 

\subsection{Phi family.}
We evaluated Microsoft’s \textbf{Phi-4}~\cite{abdin2024phi4technicalreport} (\(\sim14\text{B}\) dense decoder) and its
lightweight derivatives (\texttt{phi-4-mini-instruct} and
\texttt{phi-4-mini-reasoning}~\cite{abdin2025phi4reasoningtechnicalreport}, \(\sim3.8\text{B}\)). This family of LLMs was designed as
\enquote{small-data curriculum models} whose pre-training is dominated by synthetic textbook-style content rather than filtered web corpora.

\subsection{Falcon Family}
For completeness, we benchmarked \textbf{Falcon-10B-Instruct}~\cite{almazrouei2023falconseriesopenlanguage}, an Apache–2.0 decoder model trained on the RefinedWeb dataset and alignment-tuned with RLHF. 

\subsection{Google Flan family.}
Encoder–decoder baselines were covered by \textbf{Flan-T5-XXL}~\cite{chung2022scalinginstructionfinetunedlanguagemodels}
(\(11\text{B}\) parameters) and \textbf{Flan-UL2}~\cite{tay2023ul2unifyinglanguagelearning} (\(20\text{B}\)).
Both models extend the original T5/UL2 sequence-to-sequence architecture with instruction tuning on a curated mixture of over one thousand NLP tasks; an additional \texttt{attrscore\_flan\_t5\_xxl}~\cite{yue2023automaticevaluationattributionlarge} checkpoint was tested, which augments the T5-XXL Weights with token-level attribution heads for explanation capabilities.

\section{Participants and Methods} \label{sec:participants}
The \textsc{ClinIQLink} shared task was publicly released through the Codabench
evaluation platform\footnote{\url{https://www.codabench.org/competitions/5117/}},
with an accompanying containerized setup for local validation and submission
via Docker and Apptainer\footnote{\url{https://github.com/Brandonio-c/ClinIQLink_CodaBench_docker-setup}}. Submissions of models/systems over 10GB in size and requiring more compute than what is offered via codabench were also enabled via direct submission to organizers to be run on the University of Maryland HPC Zaratan. 
In total, 43 participants registered for the challenge during the initial
release window. The competition remains open for new submissions on Codabench for smaller models that can run via the Codabench platform. 

\subsection{Preceptor AI}

Although forty-three teams registered, only \textsc{Preceptor AI} submitted runnable systems.  
They provided three containerised runs, \texttt{v001}, \texttt{v002}, and \texttt{v003}, but discuss only \texttt{v001} in their participant paper.  

\paragraph{v001 – VeReaFine (Verifier-augmented RAG).}
\texttt{v001} is an iterative, evidence-seeking pipeline that couples a Qwen-7B-Instruct generator with a separately fine-tuned Qwen-8B medical-reasoning verifier.  
For each question the system:

\begin{enumerate}[leftmargin=*,nosep]
    \item retrieves up to 20 passages from a ColBERT~\cite{khattab2020colbertefficienteffectivepassage} + BM25~\cite{robertson2009probabilistic} hybrid index built over \textit{PubMed} abstracts and \textit{StatPearls};  
    \item drafts an answer \emph{with inline citations};  
    \item scores every generated claim with the verifier’s token-level entailment head;  
    \item if any claim falls below a 0.8 confidence threshold, expands the evidence pool and repeats steps (1)–(3) (max.\ four rounds).  
\end{enumerate}

The loop stops when all claims are verified or the round limit is reached, after which the final answer and citation list are emitted.  
This design yields strong gains on all four open-ended modalities (top-10 P75 recall) but was \emph{not} tuned for the closed-ended formats, explaining its low rank on multiple-choice and true/false items (see Table \ref{tab:leaderboard}).

\paragraph{v002 and v003.}
The team also submitted \texttt{v002} (a retrieval-free Qwen-32B classifier optimised for closed-ended questions) and an ablation run \texttt{v003}.  
Because their accompanying paper focuses on the verifier-augmented strategy, only \texttt{v001} is analysed in detail there; we include the headline numbers for all three runs in the leaderboard for completeness.

\section{Results} \label{sec:results}
\begin{figure}[ht]
  \centering
  \includegraphics[width=1\linewidth]{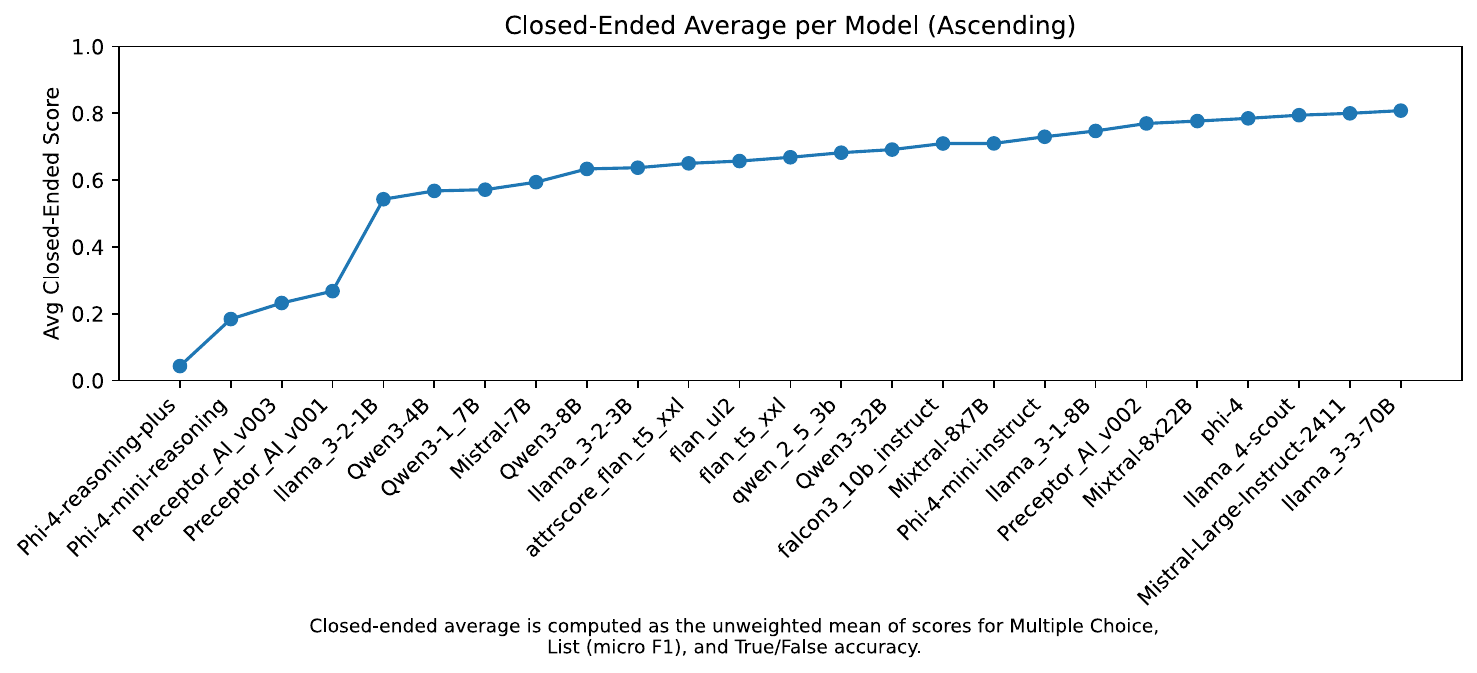}
  \caption{Average performance on closed-ended tasks (True/False accuracy, multiple-choice accuracy and list F$_1$).}
  \label{fig:avg_closed_ended}
\end{figure}

\begin{figure}[ht]
  \centering
  \includegraphics[width=1\linewidth]{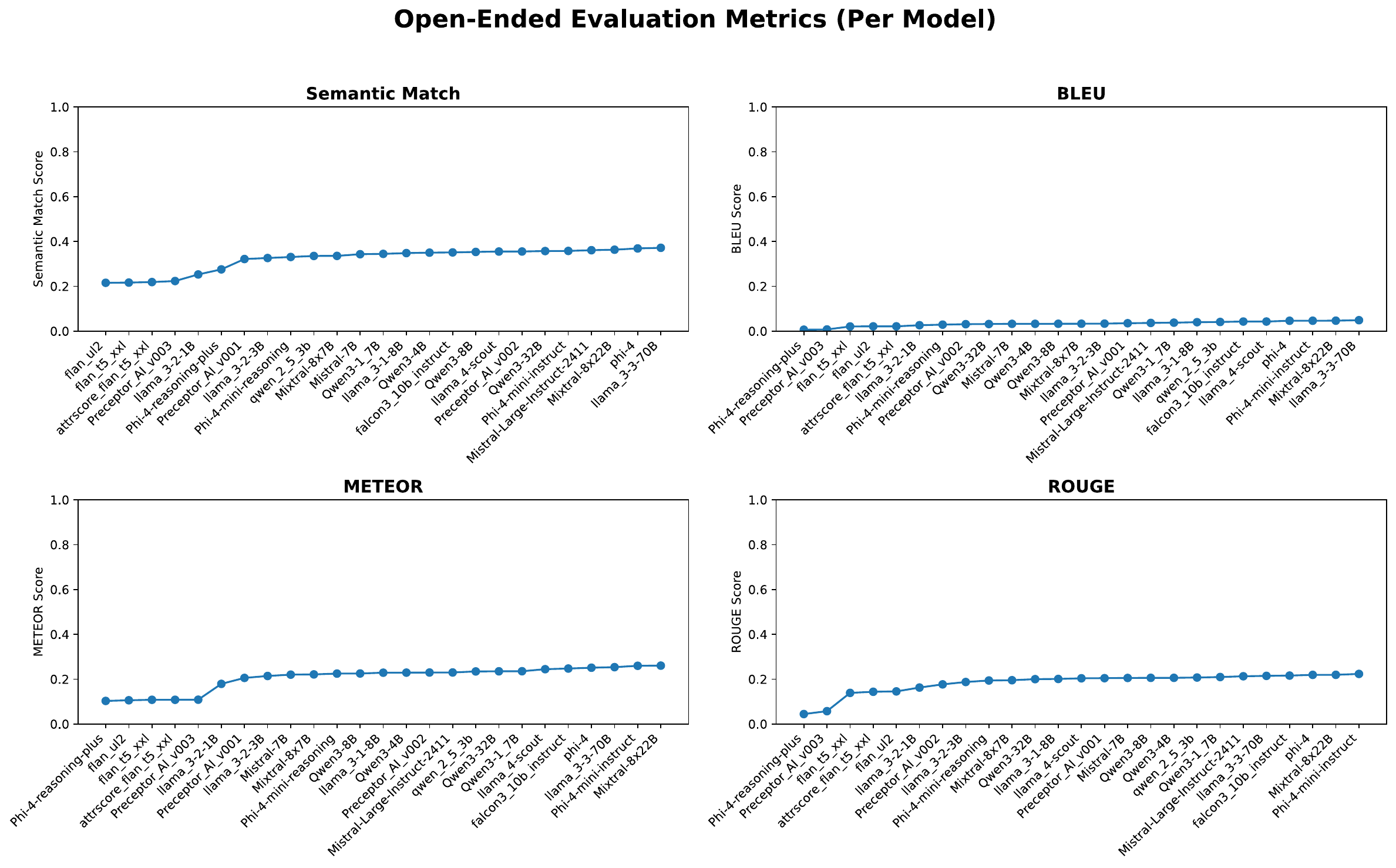}
  \caption{Distributions of individual n-gram scores (BLEU, ROUGE, METEOR) and semantic similarity for each open-ended question type.}
  \label{fig:open_ended_individual}
\end{figure}

\begin{figure}[ht]
  \centering
  \includegraphics[width=1\linewidth]{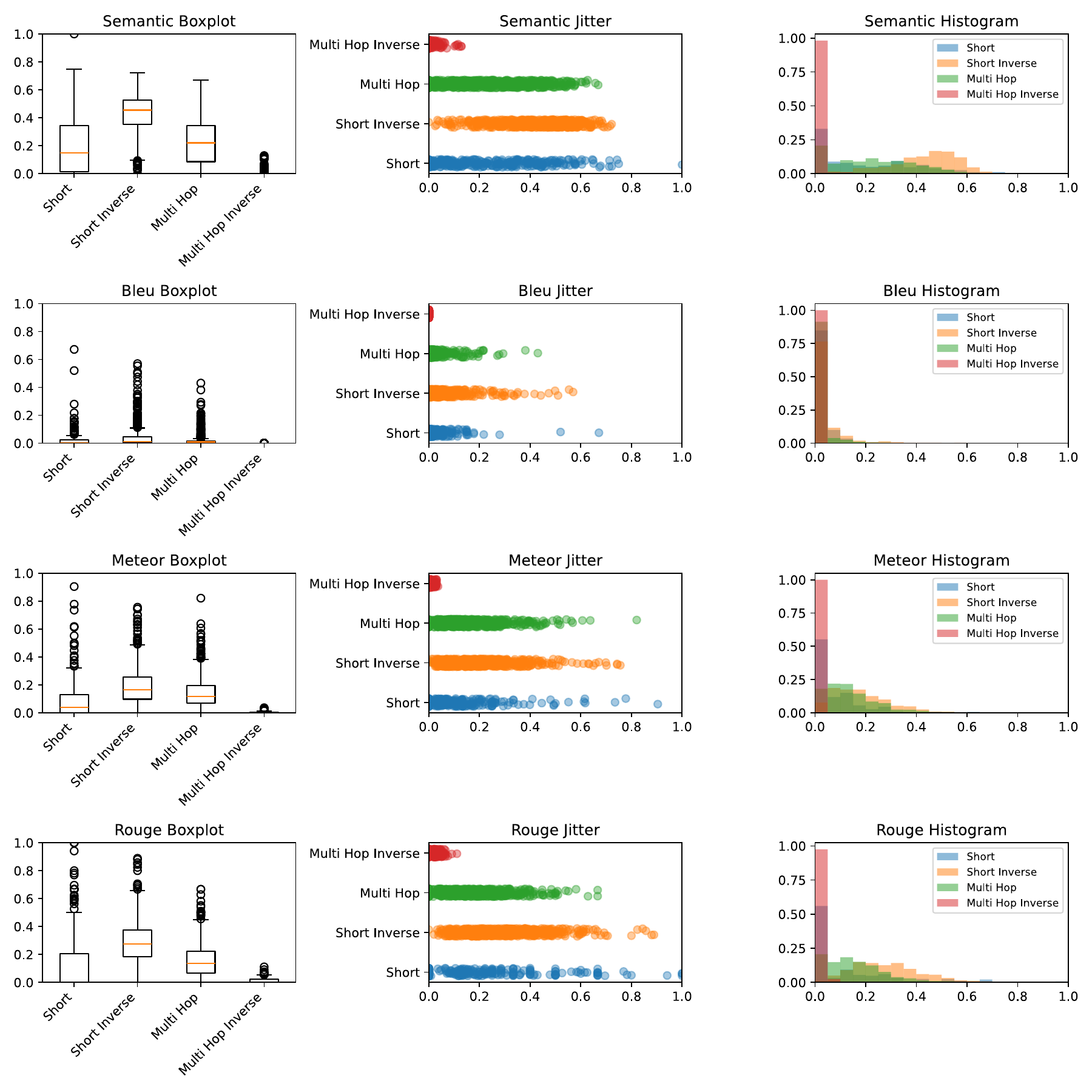}
  \caption{Comprehensive dashboard for FLAN-UL2 showing boxplots, jitter plots and histograms across semantic and n-gram metrics.}
  \label{fig:flan_ul2_dashboard}
\end{figure}

\begin{figure}[ht]
  \centering
  \includegraphics[width=1\linewidth]{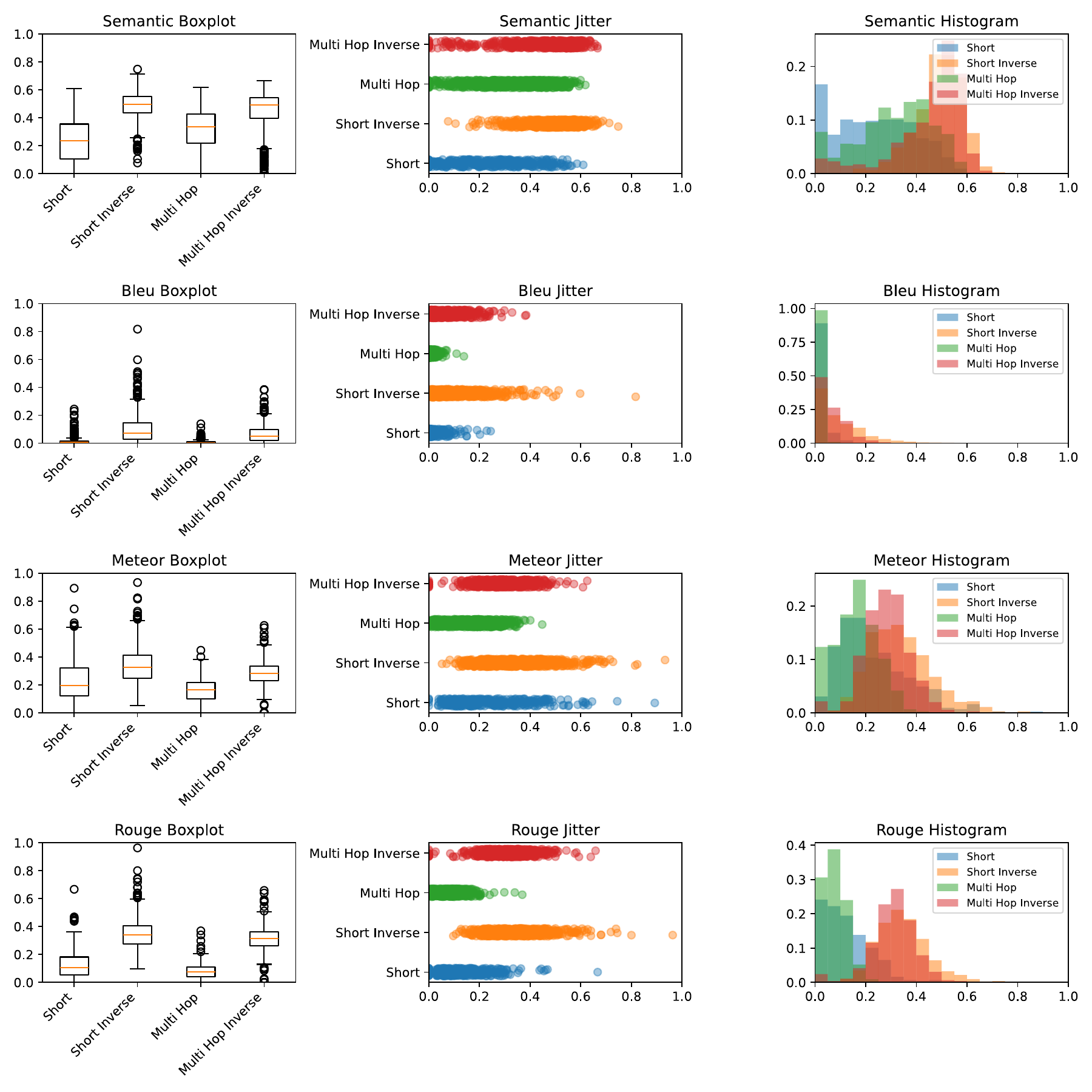}
  \caption{Comprehensive dashboard for LLaMA-70B showing boxplots, jitter plots and histograms across semantic and n-gram metrics.}
  \label{fig:llama70b_dashboard}
\end{figure}

\begin{figure}[ht]
  \centering
  \includegraphics[width=1\linewidth]{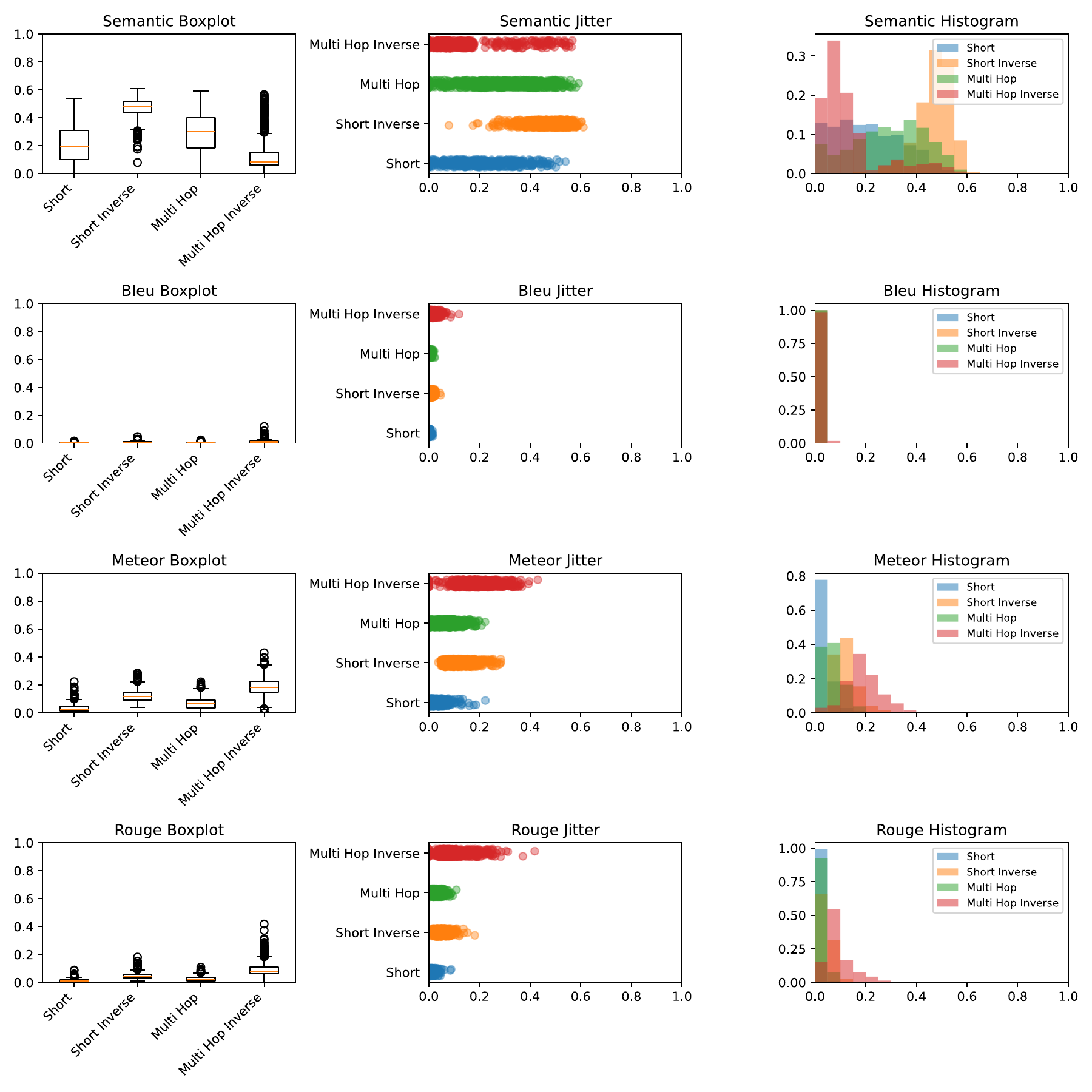}
  \caption{Comprehensive dashboard for Phi-4-reasoning-plus showing boxplots, jitter plots and histograms across semantic and n-gram metrics.}
  \label{fig:phi4_reasoning_dashboard}
\end{figure}

\begin{table*}[htbp]
  \centering
  \small            
  \caption{
    ClinIQLink leaderboard snapshot (higher is better). Models were evaluated across all seven modalities of the ClinIQLink challenge. All models were retrieved from their public \href{https://huggingface.co/}{Hugging Face} repositories, except for \texttt{Preceptor\_AI}, which is private.
    }
    
  \begin{tabular}{rlcccccccc}
    \toprule
    Rank & Model & Overall & MC Acc & TF Acc & List F1 & Short & S-Inv & MHop & MH-Inv \\
    \midrule
    1  & \texttt{llama\_3-3-70B}               & 0.541 & 0.796 & 0.822 & 0.682 & 0.235 & 0.488 & 0.313 & 0.450 \\
    2  & \texttt{Mistral-Large-Instruct-2411}  & 0.530 & 0.797 & 0.822 & 0.645 & 0.260 & 0.472 & 0.313 & 0.398 \\
    3  & \texttt{phi-4}                        & 0.528 & 0.775 & 0.790 & 0.658 & 0.229 & 0.493 & 0.311 & 0.440 \\
    4  & \texttt{llama\_4-scout}               & 0.524 & 0.776 & 0.822 & 0.652 & 0.238 & 0.492 & 0.302 & 0.388 \\
    5  & \texttt{Mixtral-8x22B}                & 0.521 & 0.752 & 0.800 & 0.643 & 0.227 & 0.491 & 0.306 & 0.428 \\
    6  & \texttt{Preceptor\_AI\_v002}          & 0.512 & 0.762 & 0.817 & 0.583 & 0.213 & 0.479 & 0.298 & 0.430 \\
    7  & \texttt{llama\_3-1-8B}                & 0.499 & 0.720 & 0.765 & 0.613 & 0.223 & 0.479 & 0.293 & 0.396 \\
    8  & \texttt{Phi-4-mini-instruct}          & 0.498 & 0.672 & 0.745 & 0.636 & 0.222 & 0.485 & 0.299 & 0.424 \\
    9  & \texttt{falcon3\_10b\_instruct}       & 0.482 & 0.673 & 0.760 & 0.538 & 0.219 & 0.487 & 0.302 & 0.396 \\
    10 & \texttt{Qwen3-32B}                    & 0.477 & 0.737 & 0.803 & 0.373 & 0.233 & 0.474 & 0.307 & 0.415 \\
    11 & \texttt{Mixtral-8x7B}                 & 0.474 & 0.656 & 0.750 & 0.570 & 0.213 & 0.472 & 0.304 & 0.353 \\
    12 & \texttt{qwen\_2\_5\_3b}               & 0.461 & 0.629 & 0.726 & 0.535 & 0.216 & 0.484 & 0.292 & 0.347 \\
    13 & \texttt{Qwen3-8B}                     & 0.454 & 0.722 & 0.748 & 0.293 & 0.223 & 0.477 & 0.316 & 0.397 \\
    14 & \texttt{llama\_3-2-3B}                & 0.436 & 0.502 & 0.733 & 0.517 & 0.200 & 0.463 & 0.271 & 0.369 \\
    15 & \texttt{Mistral-7B}                   & 0.427 & 0.425 & 0.701 & 0.491 & 0.216 & 0.483 & 0.295 & 0.378 \\
    16 & \texttt{Qwen3-4B}                     & 0.423 & 0.515 & 0.752 & 0.294 & 0.212 & 0.470 & 0.310 & 0.408 \\
    17 & \texttt{Qwen3-1\_7B}                  & 0.419 & 0.393 & 0.681 & 0.484 & 0.206 & 0.483 & 0.299 & 0.390 \\
    18 & \texttt{flan\_t5\_xxl}                & 0.390 & 0.599 & 0.705 & 0.558 & 0.220 & 0.420 & 0.220 & 0.005 \\
    19 & \texttt{flan\_ul2}                    & 0.383 & 0.567 & 0.695 & 0.556 & 0.205 & 0.430 & 0.223 & 0.003 \\
    20 & \texttt{attrscore\_flan\_t5\_xxl}     & 0.383 & 0.571 & 0.680 & 0.552 & 0.214 & 0.428 & 0.227 & 0.005 \\
    21 & \texttt{llama\_3-2-1B}                & 0.354 & 0.379 & 0.610 & 0.477 & 0.181 & 0.450 & 0.269 & 0.111 \\
    22 & \texttt{Preceptor\_AI\_v001}          & 0.295 & 0.047 & 0.713 & 0.021 & 0.163 & 0.482 & 0.277 & 0.363 \\
    23 & \texttt{Phi-4-mini-reasoning}         & 0.249 & 0.095 & 0.068 & 0.256 & 0.196 & 0.456 & 0.281 & 0.389 \\
    24 & \texttt{Preceptor\_AI\_v003}          & 0.221 & 0.000 & 0.581 & 0.074 & 0.111 & 0.286 & 0.233 & 0.263 \\
    25 & \texttt{Phi-4-reasoning-plus}         & 0.167 & 0.000 & 0.000 & 0.070 & 0.206 & 0.470 & 0.290 & 0.135 \\
    \bottomrule
  \end{tabular}
  \label{tab:leaderboard}
\end{table*}

Figure~\ref{fig:avg_closed_ended} summarised mean performance on the three closed-ended tasks.  
The spread between True/False, multiple-choice and list accuracy was modest, indicating that the leading models handled discrete answer formats with broadly comparable competence.

Open-ended behaviour was more nuanced.  
The per-task distributions in Figure~\ref{fig:open_ended_individual} showed markedly heavier tails for semantic-similarity than for surface n-gram metrics, confirming that several systems produced answers that were lexically novel yet semantically similar.  
This pattern was especially pronounced for the \texttt{multi-hop} and \texttt{multi-hop inverse} questions, where BLEU occasionally under-estimated quality relative to the embedding‐based score.

To illustrate model-specific traits, Figures~\ref{fig:flan_ul2_dashboard}–\ref{fig:phi4_reasoning_dashboard} present the full metric dashboards for three representative baselines.  
The \textsc{FLAN-UL2} run exhibited tight clustering around mid-range similarity values and an extreme outlier for the multi-hop inverse modality.

\textsc{LLaMA-3 70B} displayed a broader inter-quartile range on semantic scores but maintained competitive n-gram fidelity, suggesting flexible paraphrasing capabilities.  

Similarly, the \textsc{Phi-4-Reasoning-Plus} submission produced a long tail of semantically similar scores when evaluated with the CLinIQLInk semantic similarity metric, but low scoring across all the n-gram scoring metrics utilised; further inspection of the model responses revealed that, despite using the prescribed stop tokens and output template, the model frequently emitted extensive chain-of-thought traces capped by an ambiguous or missing \enquote{final answer} cue.  
Our automated evaluation script extracted only the required answers utilising pre-determined queues (i.e. the prompt templates used explicitly constrained models to provide list-type responses as comma-separated lists, etc.) and as such, the digressions observed from the Phi-4-Reasoning-Plus (amongst others) translated into poor task compliance rather than genuine comprehension deficits.

A consolidated leaderboard is provided in Table~\ref{tab:leaderboard}.  
The ranking served solely as an empirical reference from the evaluation metrics gathered from task 1 automated evaluation script.

\section{Discussion} \label{sec:discussion}
\subsection{Closed-ended tasks (Figure~\ref{fig:avg_closed_ended}).}
Table~\ref{tab:leaderboard} confirms what is visually apparent in the right-hand side of Figure~\ref{fig:avg_closed_ended}, which is that single-labelled questions (e.g., t/f, MC, etc.)  are close to saturation for modern LLMs.  
The top five systems tested (\texttt{llama\_3-3-70B}, \texttt{Mistral-Large}, \texttt{phi-4}, \texttt{llama\_4-scout}, and \texttt{Mixtral-8x22B}) all scored between 0.75 – 0.80 on \textbf{multiple-choice} and 0.79 – 0.82 on \textbf{True/False}.  
By contrast, \textbf{list} questions remained challenging with macro–micro~F$_1$ not exceeding~0.68. List answers required both recognition of all correct options \emph{and} rejection of distractors, and as such, the metric penalised even minor hallucinations; consequently, models whose generation style tended to \enquote{hedge} with extra choices (e.g. \texttt{falcon3\_10b\_instruct}) underperformed relative to their multiple-choice score.  The tight inter-quartile ranges on True/False and multiple-choice further suggest that most contemporary LLMs share a common ceiling on purely factual one-shot classification, leaving little room for architectural distinctions to distinguish in these settings.

\subsection{Aggregate open-ended behavior} Figure~\ref{fig:open_ended_individual} shows that, 
across all models evaluated, the \texttt{short inverse} distributions peaked around 0.50 semantic similarity, while the forward \texttt{short} items clustered near 0.25, indicating that simply \emph{critiquing} providing an answer was easier than generating an answer from scratch.  
The gap between semantic and n-gram scores widens for larger checkpoints. Mixtral-8×22B and LLaMA 3.3 70B frequently achieved high semantic similarity scores (above 0.60) despite very low BLEU scores (below 0.1), indicating that their correct answers were often paraphrased rather than copied verbatim, supporting the long-tailed distribution of paraphrastic responses seen in Figure~\ref{fig:open_ended_individual}. 
Inspection of the model responses for multi-hop inverse QA types also revealed answers that often diagnosed the wrong knowledge hop step, which in turn attracted the multiplicative penalties. Traditional n-gram metrics failed to flag these omissions, underscoring the necessity of the custom semantic evaluation platform.

\subsection{Model-specific open-ended evaluations} Figures~\ref{fig:flan_ul2_dashboard}–\ref{fig:phi4_reasoning_dashboard} illustrate how  aggregate patterns materialised at the system level. The model-specific open-ended evaluations are shown for only the highest performing model across the board (llama-3.3 70B and the lowest performing model for closed and open-ended metrics (Phi-4-reasoning and FLAN-UL2, respectively).

\begin{itemize}
\item \textbf{FLAN-UL2} 
Figure \ref{fig:flan_ul2_dashboard} reveals that FLAN-UL2’s outputs cluster tightly between 0.20 and 0.60 for the three forward-facing open-ended tasks, yet its \texttt{multi-hop inverse} scores collapse toward the origin on \emph{all four} axes—semantic similarity, BLEU, ROUGE, and METEOR rarely rise above 0.05. The dashboard traces that floor effect to the model’s habit of supplying only a step label (e.g., \enquote{Step 5}) with no explanatory text, which earns minimal credit under the step-penalised rubric. Elsewhere, list questions are answered with bare option letters (e.g. \enquote{B, C, D}), boosting recall but cutting precision to roughly 0.33–0.50, while short prompts receive one or two-word noun phrases, driving n-gram metrics to zero even when the semantics are acceptable. These abrupt, template-bound behaviours keep variance low and prevent catastrophic errors, but they also cap the weighted open-ended average at 0.14 and hold FLAN-UL2 in 18\textsuperscript{th} place despite competent closed-ended performance.

\item \textbf{LLaMA-3-3-70B} Figure \ref{fig:llama70b_dashboard} shows that LLaMA-3 70 B’s open-ended answers cluster in the mid-range for every metric, not at the extremes. Its \emph{semantic-similarity} box-plot sits roughly between 0.35 and 0.55, with whiskers reaching only the mid-0.70s; BLEU, ROUGE, and METEOR centre much lower (BLEU’s median is barely above 0.04, ROUGE around 0.20, METEOR around 0.25). The small band of higher-value semantic outliers (around 0.65–0.75) is confined to \texttt{short inverse} and \texttt{multi-hop inverse} items in which the model repeated key medical terms but reordered the surrounding sequence of words, so n-gram overlap stayed muted. Conversely, many \texttt{short} replies are abrupt noun-phrases, depressing all four metrics and keeping the inter-quartile ranges tight.

\item \textbf{Phi-4-Reasoning-Plus} (Figure~\ref{fig:phi4_reasoning_dashboard}).  
The cloud at the extreme lower-left of the dashboard mirrors the 624 malformed list entries and 813 invalid True/False lines produced by this model.  Extensive \enquote{chain-of-thought} preambles obscured the required delimiters, so the automated evaluation script extracted empty or partial lines.  BLEU/ROUGE medians (around 0.04) remained higher than the semantic median (around 0.02) because the responses still shared surface n-grams with the references.
\end{itemize}

\subsection{Cross-metric contrasts.}
\begin{enumerate}
  \item The ClinIQLink Semantic similarity metric displayed higher variance than any n-gram metric across every model dashboard, reflecting sensitivity to both omissions \emph{and} verbose digressions.
  
  \item The gap between ClinIQLink Semantic similarity metric and BLEU was inversely correlated with parameter count; smaller Qwen checkpoints recycled reference wording, whereas 70-B LLaMAs paraphrased aggressively.
  
  \item \texttt{Multi hop inverse} was the most discriminative sub-task; its step-penalty compressed medians for every system (lowest boxes in Figure~\ref{fig:open_ended_individual}), frequently reshuffling neighbouring ranks in Table~\ref{tab:leaderboard}.
  
\end{enumerate}

\subsection{Findings}

\begin{itemize}

  \item High closed-ended scores hide residual hallucinations. Even with a vocabulary capped at just true/false or four choice-letters, every model occasionally invented an out-of-range option, proving that 0.75–0.82 headline accuracies do not equal flawless control.

  \item List questions are the singular closed-ended format that is still able to effectively discriminate model effectiveness because they demand selecting all true items while rejecting distractors, and as such, macro–micro F$_1$ was found to be spread from 0.30 to 0.68.  Those wider answer sets surface the hallucinated extras that multiple-choice and true/false conceal.

  \item \enquote{Critique} is easier than \enquote{generate}. Across the board, \texttt{short inverse} prompts (spot the error) cluster around 0.50 semantic similarity which is roughly double the median for forward \texttt{short} prompts that require composing a fresh answer.

  \item Multi-hop-inverse is the most discriminative open-ended task. Its step-distance penalty drags every model’s median to the bottom of Figure~\ref{fig:open_ended_individual}, reshuffling several adjacent leaderboard positions and exposing brittle reasoning chains.

  \item Embedding-level similarity scores for LLM evaluation tasks are now required as the minimum standard. High-ranked systems such as Mixtral 22B and LLaMA-3.3 70B often score > 0.60 on the semantic metric while BLEU, ROUGE and METEOR sit < 0.05, confirming that lexically novel yet faithful paraphrases fool token-overlap measures. Conversely, runs that recycle reference text earn decent n-gram scores but remain low on the embedding metric, demonstrating that overlap alone no longer tracks answer fidelity.

  \item Note on open-ended evaluation challenges. Despite impressive progress in embedding-based metrics (e.g., BERTScore, Sentence-Mover, BLEURT, COMET etc.) and NLI-based metrics (e.g., MENLI, UniEval), no single method can yet (a) decide with high confidence that two free-form LLM responses convey the same meaning, while also (b) grounding that decision in consistent entity and relation alignment across passages. Embedding similarity captures distributional closeness but is blind to logical entailment; NLI classifiers reason over sentence-level entailment yet lack explicit entity grounding and scale poorly beyond short contexts, and recent surveys and benchmark studies conclude that integrating these complementary views into a robust, scalable metric remains an unsolved problem and a key direction for future work \cite{Ito2025, Croxford2025}.
  
\end{itemize}

\section{Conclusion} \label{sec:conclusion}
The \textsc{ClinIQLink} evaluation shows that modern LLMs reach impressive headline scores on tightly constrained \textit{True/False} and single-letter \textit{multiple-choice} items, yet every model evaluated still sporadically produces out-of-vocabulary or otherwise invalid answers; unordered \textit{list} questions, with their wider response space, remain the only closed-ended format able to expose this fragility.  On open-ended tasks, embedding-based semantic similarity distinguishes genuinely informative paraphrases from superficial n-gram overlap. Conventional n-gram indices systematically mis-score open responses, rewarding superficial token overlap while penalising lexically novel yet factually correct paraphrases; embedding-based similarity aligns far more closely with clinical accuracy and, through the step-penalised \textit{multi-hop-inverse} task, reveals brittle reasoning chains. More work is required to produce an effective semantic similarity scoring metric with explicit reasoning validation into a composite metric that more rigorously captures factuality, logical coherence, entity relationship framing, and schema compliance. To support this goal, future iterations of \textsc{ClinIQLink} will link each question–answer pair to a machine-readable knowledge graph for graph-based verification of multi-step rationales and will introduce multimodal variants that couple text queries with images, thereby challenging models to ground their answers in heterogeneous clinical evidence.

\section*{Acknowledgments}
This research was supported in part by the Division of Intramural Research (DIR) of the National Library of Medicine (NLM), National Institutes of Health. This work utilized the computational resources of the NIH HPC Biowulf cluster (\url{https://hpc.nih.gov}) and the University of Maryland’s Zaratan cluster (\url{https://hpcc.umd.edu/hpcc/zaratan.html}). This research was also partly supported by the U.S. Fulbright program, enabling international collaborative efforts. 

\bibliography{custom}

\onecolumn
\clearpage

\end{document}